\documentclass{llncs}
\usepackage{hyperref}
\usepackage{breakurl}
\usepackage{makeidx}  
\usepackage[sectionbib,numbers,sort&compress]{natbib}
\usepackage{footnote}
\usepackage{graphicx}
\usepackage{wrapfig}
\begin{document}
\pagestyle{headings}  
\mainmatter              
\title{Challenges in Representation Learning: A report on three machine learning contests}
\titlerunning{Challenges in Representation Learning}  
%
\author{
Ian J. Goodfellow\inst{1} \and Dumitru Erhan\inst{2} \and
Pierre Luc Carrier \and Aaron Courville \and Mehdi Mirza \and
Ben Hamner \and Will Cukierski \and
Yichuan Tang \and David Thaler
\and Dong-Hyun Lee \and Yingbo Zhou \and Chetan Ramaiah \and Fangxiang Feng
\and Ruifan Li \and Xiaojie Wang
\and Dimitris Athanasakis \and John Shawe-Taylor \and Maxim Milakov \and John Park
\and Radu Ionescu \and Marius Popescu \and Cristian Grozea
\and James Bergstra \and Jingjing Xie
\and Lukasz Romaszko
\and Bing Xu \and Zhang Chuang
\and Yoshua Bengio
}
\authorrunning{Ian Goodfellow et al.} 
\institute{Universit\'{e} de Montr\'{e}al, Montr\'{e}al QC H3T 1N8, Canada,\\
\email{goodfeli@iro.umontreal.ca}
\and
Google,
Venice, CA 90291, USA,\\
\email{dumitru@google.com}}

\maketitle              

\begin{abstract}
The ICML 2013 Workshop on Challenges in Representation Learning\footnote{\url{http://deeplearning.net/icml2013-workshop-competition}}
focused on three challenges: the black box learning challenge, the facial expression
recognition challenge, and the multimodal learning challenge. We describe the
datasets created for these challenges and summarize the results of the
competitions. We provide suggestions for organizers of future challenges and
some comments on what kind of knowledge can be gained from machine learning
competitions.
\keywords{representation learning, competition, dataset}
\end{abstract}
\section{Introduction}

This paper describes three machine learning contests that were held as part
of the ICML workshop ``Challenges in Representation Learning.'' The purpose
of the workshop, organized by Ian Goodfellow, Dumitru Erhan, and Yoshua Bengio,
was to explore the latest developments in representation learning, with a
special emphasis on testing the capabilities of current representation learning
algorithms (See \citep{Bengio+Courville+Vincent-arxiv2012} for a recent review)
and pushing the field towards new developments via these contests.
%
%
%
Ben Hamner and Will Cukierski handled all issues related to Kaggle hosting
and ensured that the contests ran smoothly.
Google provided prizes for all three contests. The winner of each contest received \$350 while
the runner-up received \$150.
%
A diverse range of competitors spanning academia, industry, and amateur machine learning
provided excellent solutions to all three problems. In this paper, we summarize their solutions,
and discuss what we can learn from them.

\section{The black box learning challenge}
\begin{wrapfigure}{l}{0.5\textwidth}
\includegraphics[width=0.4\textwidth]{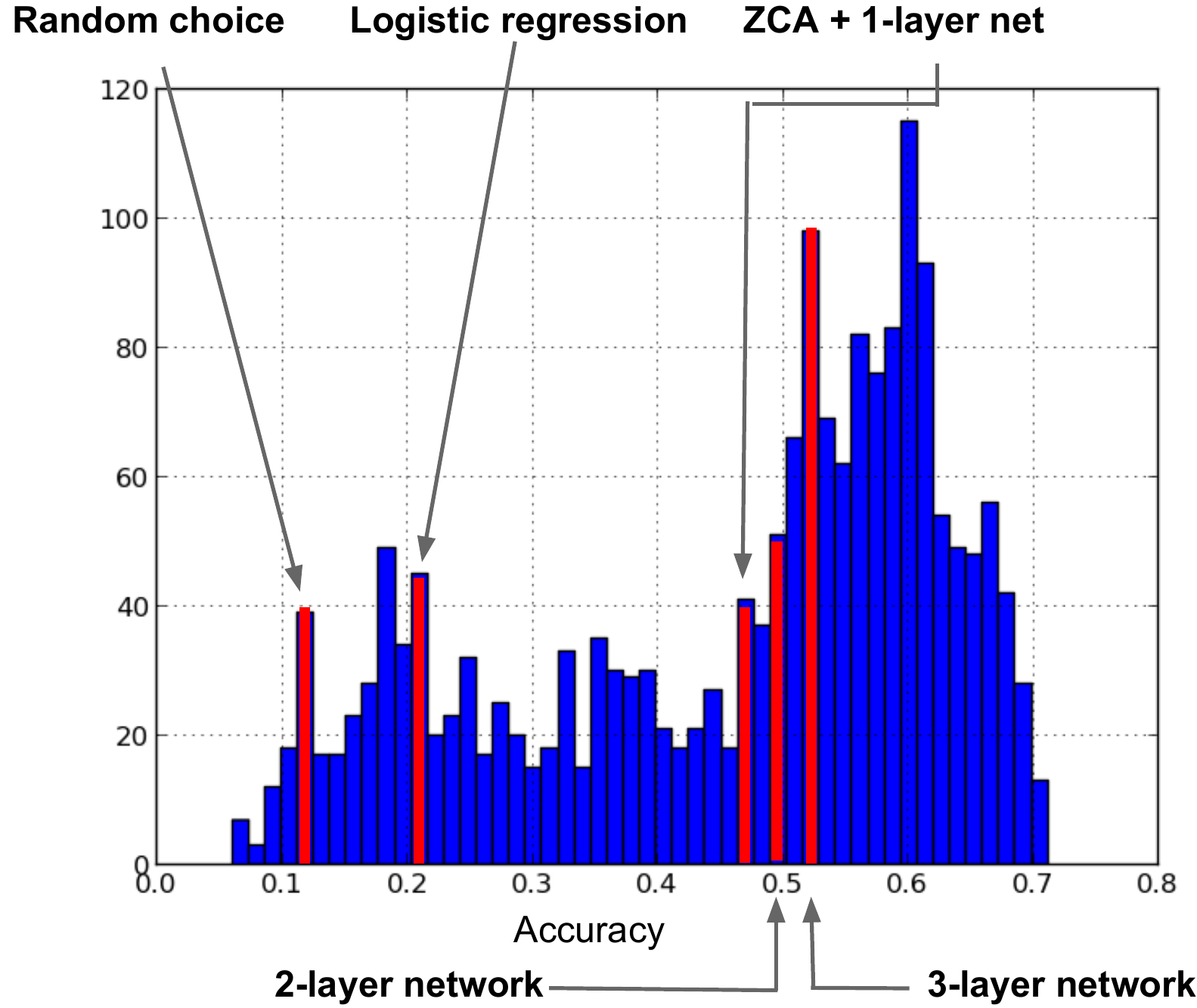}
\caption{\small{Histogram of accuracies obtained by different submissions on the BBL-2013 dataset.
Organizer-provided baselines shown in red.}
\label{bbl_hist}}
\end{wrapfigure}

The black box learning challenge\footnote{\tiny{\url{http://www.kaggle.com/c/challenges-in-representation-learning-the-black-box-learning-challenge}}}
was designed with two goals in mind. First, the data was obfuscated, so that competitors could not use human-in-the-loop techniques like
visualizing filters to guide algorithmic development. A common criticism of deep learning is that it is an art requiring an expert practitioner.
By keeping the domain of the data secret, this contest reduced the usefulness of the human practitioner. 
This idea was similar to a recent DARPA-organized unsupervised and transfer learning challenge \citep{Guyon-UTLC-ijcnn2011} which used
obfuscated data and required submission of a representation of the data that would then be used on the competition server to train a very weak
classifier. In this contest, we allowed competitors to use any method; using representation learning was not a requirement.
The second goal of this contest was to test the ability of algorithms to benefit from extra unsupervised data. To this end, we provided only
very few labeled examples.


This contest introduced the Black Box Learning 2013 (BBL-2013) dataset.
The scripts needed to re-generate it are available for download\footnote{\tiny{\url{http://www-etud.iro.umontreal.ca/~goodfeli/bbl2013.html}}}.
The dataset is an obfuscated subset of the second (MNIST-like) format of the Street View House Numbers dataset\citep{Netzer-wkshp-2011}.
Dumitru Erhan created the dataset. The original data contained 3,072 features (pixels) which he projected down to 1875 by multiplication
by a random matrix. He also removed one class (the ``4''s). These measures obfuscated the data so competitors did not know what task they
were solving. The organizers did not reveal the source of the dataset until after the contest was over. To make the challenge emphasize
semi-supervised learning, only 1,000 labeled examples were kept for training. Another 5,000 were used for the public leaderboard. For these
examples, the labels are not provided to the competitors, but the features are. Each team may upload predictions for these examples twice
per day. The resulting accuracy is published publicy. The public test set is thus a sort of validation set, but also gives one's competitors
information. Another 5,000 examples were used for the private test set. The features for these examples are given to the competitors as well,
but only the contest administrators see the accuracy on them until after the contest has ended. The private test set is used to determine
the winner of the contest. We also provided 130,000 unlabeled examples drawn from a set specified to be ``less difficult'' by the creators of
SVHN. 


218 teams submitted 1963 entries to the contest. 75 teams beat the best baseline (a 3-layer MLP) provided by the organizers.
See Fig. \ref{bbl_hist} for a histogram of all the teams' performance.
David Thaler won the contest with an accuracy of 70.22\% using blending of three models that used sparse filtering\citep{Ngiam11} for feature learning,
random forests for feature selection \citep{Breiman01}, and
support vector machines\citep{Cortes95} for classification. Other competitors such as Lukasz Romaszko \citep{Romaszko-wkshp-2013} also obtained
very competitive results with sparse filtering. This was an interesting outcome because sparse filtering has usually been perceived as
an inexpensive and simple method that gives good but not optimal results. David Thaler and Lukasz Romaszko both observed that
learning the sparse filtering features on the combination of the labeled and unlabeled data worked {\em worse} than learning
the features on just the labeled data. This may be because the labeled data was drawn from the more difficult portion of the
SVHN dataset. Dong-Hyun Lee \citep{Lee-wkshp-2013} finished second in the contest, having independently rediscovered entropy regularization
\citep{GrandvaletY2005}.
This very simple means of semi-supervised learning proved surprisingly effective and merits more attention.
In third place, Dimitris Athanasakis and John Shawe-Taylor developed a new feature section / combination mechanism combined with MKL.
Other top scorers included Jingjing Xie, Bing Xu and Zhang Chuang, that developed ensemble voting techniques for use with
denoising autoencoders \citep{VincentPLarochelleH2008-small} and maxout networks \citep{Goodfellow+al-ICML2013-apj}.

A recent trend in deep learning has been to forego unsupervised learning entirely following recent improvements to discriminative training. 
This is probably a result of most datasets having several labeled examples. In this contest, with only 1,000 labeled training
examples, most of the top scorers still needed to make use of the unlabeled data in some way. 

\section{The facial expression recognition challenge}

In the facial expression recognition challenge\footnote{\tiny{\url{http://www.kaggle.com/c/challenges-in-representation-learning-facial-expression-recognition-challenge}}}
we invited competitors to design the best system for recognizing which emotion is being expressed in a photo
of a human face. In this contest, we wanted to compare methods on a task that is well studied but using a completely new dataset.
This avoids issues of overfitting to the test set of a repeatedly used benchmark dataset. One reason to hold such a contest is that
it allows us to compare feature learning methods to hand-engineered features in as fair a manner as possible.


This contest introduced the Facial Expression Recognition 2013 (FER-2013) dataset.
It is available for download\footnote{\tiny{\url{http://www-etud.iro.umontreal.ca/~goodfeli/fer2013.html}}}.
FER-2013 was created by Pierre Luc Carrier and Aaron Courville. It is part of a larger ongoing project.
The dataset was created using the Google image search API to search for
images of faces that match a set of 184 emotion-related keywords like ``blissful'', ``enraged,'' etc.
These keywords were combined with words related to gender, age or ethnicity,
to obtain nearly 600 strings which were used as facial image search queries. The first 1000 images returned for
each query were kept for the next stage of processing. 
OpenCV face recognition was used to obtain bounding boxes around each face in
the collected images.
Human labelers than rejected incorrectly labeled images, corrected the cropping if necessary, and filtered out some duplicate images.
Approved, cropped images were then resized to 48x48 pixels and converted to grayscale. Mehdi Mirza and Ian Goodfellow prepared
a subset of the images for this contest, and mapped the fine-grained emotion keywords into the same seven broad categories
used in the Toronto Face Database \citep{Susskind2010}. The resulting dataset contains 35887 images, with 4953 ``Anger'' images,
547 ``Disgust'' images, 5121 ``Fear'' images, 8989 ``Happiness'' images, 6077 ``Sadness'' images, 4002 ``Surprise'' images, and
6198 ``Neutral'' images.


Ian Goodfellow
performed some small-scale experiments to estimate the human performance on this task.
He collected 1500 images of members of the LISA lab acting out the seven facial expressions.
This dataset contains no label noise per se, though poor acting abilities mean that the Bayes
rate could be quite high. On this dataset, human accuracy was 68$\pm$5\%. FER-2013 could
theoretical suffer from label errors due to the way it was collected, but Ian Goodfellow found
that human accuracy on FER-2013 was 65$\pm$5\%. While there may be label errors, they do
not make the task significantly harder, at least not for a human. James Bergstra also determined
the best performance of a ``null'' model, consisting of a convolutional network with no learning
except in the final classifier layer. Using the TPE hyperparameter optimization algorithm, he found
that the best such convolutional network obtains an accuracy of 60\%. Using an ensemble of such models,
he obtained an accuracy of 65.5\%. See \citep{Bergstra-wkshp-2013} for details.

56 teams submitted on the final dataset. Of these, four beat the best ``null'' ensemble model (which was not presented
until after the contest was over--many more teams beat the simpler baselines provided by the organizers).
Their scores are presented in Table \ref{fer-leaderboard}. The top three teams all used convolutional neural networks \citep{Fukushima80}
trained discriminatively with image transformations. The winner, Yichuan Tang, used the primal objective of an SVM
as the loss function for training. This loss function has been applied to neural networks before, but he additionally used the L2-SVM
loss function, a new development that gave great results on the contest dataset and others.

One of the questions we hoped to answer in this workshop is whether or not feature learning algorithms are ahead of other methods.
Radu Ionescu, Marius Popescu, and Cristian Grozea provided the strongest submission that did not use feature learning. Their approach
used SIFT \citep{Lowe99} and MKL. This approach put their performance close to that of Maxim Milakov, who submitted the third best convolutional network.
These results suggest that convolutional networks are indeed capable of outperforming hand-designed features, but the difference in accuracy
is not extreme. It's unclear whether the performance of the best deep network has reached the Bayes rate on this task or not.

\begin{savenotes}
\begin{table}[t]
\tiny
\caption{Private test set accuracy on FER-13}
\label{fer-leaderboard}
\vskip 0.15in
\begin{center}
\begin{small}
\begin{sc}
\begin{tabular}{p{5cm}|p{5cm}|c}
Team & Members & Accuracy \\
\hline
RBM \citep{Tang-wkshp-2013} & Yichuan Tang & 71.162\%\\
Unsupervised & Yingbo Zhou, Chetan Ramaiah & 69.267\%\\
Maxim Milakov\footnote{\url{http://nnforge.org}} & Maxim Milakov & 68.821\%\\
Radu + Marius + Cristi \citep{Ionescu-wkshp-2013} & Radu Ionescu, Marius Popescu, Cristian Grozea & 67.484\%\\
\end{tabular}
\end{sc}
\end{small}
\end{center}
\vskip -0.1in
\end{table}
\end{savenotes}

\section{The multimodal learning challenge}

The multimodal learning challenge\footnote{\tiny{\url{http://www.kaggle.com/c/challenges-in-representation-learning-multi-modal-learning}}}
was intended to spur development of algorithms that discover a unified
semantic representation of examples that have more than one input representation. In this case,
the two input modalities were images and text.


Competitors were advised to use the small ESP game dataset \citep{vonAhn2004} for training data,
but all public sources of training data were allowed. The small ESP game dataset consists of 100,000
images of varying sizes that were annotated by players of an online game. Each image is tagged with
on average 14 words, with a vocabulary of over 4,000 words.

In order to provide a new test set, Ian Goodfellow manually labeled 1,000 images obtained by Google
image search queries for some of the most commonly used words in the small ESP game dataset. The
labels were intended to resemble those in the training set. For example, they include incorrect
spellings that were common in the training set. This dataset is available for download\footnote{\tiny{\url{http://www-etud.iro.umontreal.ca/~goodfeli/mlc2013.html}}}.


Kaggle does not yet provide the kinds of evaluation metrics typically used for multimodal learning, so
the organizers devised a multimodal classification task. Each test image would be accompanied by two labels
from the test set, with the classification task being to report which of the two labels is
correct. Unfortunately, because this is a matching task, it proved too easy to yield interesting
machine learning results. Yichuan Tang found that a base classifier with low accuracy could be coupled
with the Hungarian algorithm to compute the optimal matching. The optimal matching constructed in
this way obtained 100\% accuracy.
The contest ended in a three-way tie with 100\% test accuracy. The winners were ``RBM'' (Yichuan Tang),
``MMDL'' \citep{Feng-wkshp-2013} (Fangxiang Feng, Ruifan Li, and Xiaojie Wang), and ``AlbinoSnowman'' (John Park). RBM won the
tie by submitting the first perfect solution. The tie between MMDL and AlbinoSnowman was broken
because MMDL submitted a model file for verification and AlbinoSnowman did not.

If a similar contest is organized in the future, we recommend labeling twice as many test images as
are needed, then discarding half of the images and using their labels as the incorrect label for the
remaining labels. This removes the matching aspect of the problem and forces the classifier to label
each image independently.

\section{Advice to contest organizers}

Organizing a contest requires a significant amount of work from all parties involved.
We offer some suggestions for running a succesful contest:

{\bf Allocation of time:} Budget time for the following tasks: {\em Before the contest launches:} 
Creation of new datasets, verification that state of the art algorithms perform well but have room
for improvement on the dataset, preparation of baseline solutions, design of rules for the contest.
{\em During the contest: } Fielding questions (on contest rules, how to use the contest website, etc.),
resolving portability issues with contest baselines. {\em After the contest} Verification of the winners'
submissions, distributing the private test data, preparing presentations and papers about the contest.

{\bf Designing rules:} Some things to consider:
Should ``transductive'' methods that are allowed to observe all test set inputs be allowed? Are contestants prohibited from
labeling the public leaderboard test data and training or cross-validating with it? What about training with outside data, or scraping the web for higher resolution versions of input images? How will you enforce the rules?
Datasets that humans can label present many difficulties. Remember that you need to prevent not just training on the test set, but also selecting hyperparameters on it.
The best way to do this is to require all entrants to upload their trained models at the end of the contest. The organizers
release the test set only after all models are frozen. Entrants then run their submission on the test set and upload the predictions
. The organizers then verify that the winning submissions' predictions were indeed generated by the previously uploaded model.
Using this system is a powerful deterrent to cheating. In order to run the contest smoothly, it is important to plan these
measures in advance and put them in the rules from the start. We initially had fewer cheating deterrants in place, expecting
only a small number of competitors from the academic deep learning community, but within days of launching the contest someone
had already hand-labeled the entire public test set for the multimodal learning contest. Note that contestants are interested in obtaining a high
rank on the leaderboard even if they do not win a prize (on Kaggle, one can earn ``Kaggle points'' for placing in the top 10\% or 25\% of a contest).
It's important to reserve the right to verify all submissions and remove leaderboard entries that can't be verified.

{\bf Difficulty and participation rate:} Err on the side of making the contest too hard rather than too easy.
We erred on the side of making the contests easy, in order to increase participation, and this made the multimodal contest
too easy to be interesting. While past workshop-based contests have had a low participation rate (example: 4
teams in the NIPS 2011 transfer learning challenge \citep{NipsWorkshop11Hierarchical,Goodfellow+al-ICML2012})
the participation rate problem can be completely solved by hosting the contest on Kaggle. Even our least popular
challenge had 26 teams.

{\bf Organize multiple contests simultaneously.} The marginal cost of running a second
or third contest is low compared to the fixed cost of launching one contest, and the additional contests provide
some insurance that you will obtain interesting results even if one contest turns out to be poorly devised. 

{\bf Provide baselines and a leaderboard.} Baselines boost participation since entrants don't
need to write boilerplate code to load the data, etc.

\section{Discussion and conclusion}

Competitions offer a different and important viewpoint on machine learning algorithms than research papers do.
Research papers are expected to be extremely novel. When writing research papers, the most talented machine learning practitioners
focus their skills on tuning methods that they themselves invented. Contests offer the opportunity to see what happens when
a different incentive structure is applied: skilled practitioners use whatever means they think will help them win, regardless of
how novel the method is or whether they invented it.
The use of a completely new test set also makes the results of the contest a more realistic evaluation of generalization error.
When interpreting the results of a contest, it is important to remember that
a contest is not a controlled experiment complete with statistical analysis. However, contests can serve to refocus our attention
on algorithms that perform well, but may not otherwise receive their due attention in the research community. This year's contest
highlighted the performance of SVM loss functions, sparse filtering, and entropy regularization. We hope these results help machine
learning practitioners improve the performance of their algorithms, and that future contest organizers are able to use this report
to plan more contests that highlight more effective algorithms.

\bibliography{wtf_apj,strings,strings-shorter,ml,aigaion,aigaion-short}
\bibliographystyle{unsrtnat}

\end{document}